\pdfoutput=1

\documentclass[11pt]{article}

\usepackage{acl}

\usepackage{times}
\usepackage{latexsym}

\usepackage[T1]{fontenc}

\usepackage[utf8]{inputenc}

\usepackage{microtype}

\usepackage{graphicx}
\usepackage{xcolor}
\usepackage{multirow}
\usepackage{booktabs} 
\usepackage{amsmath}
\usepackage{amssymb}
\usepackage{fontawesome}
\usepackage{linguex}
\usepackage{tablefootnote}
\usepackage{tikz-dependency}
\usepackage{amsmath}
\usepackage{color, colortbl}

\title{What's Hard in English RST Parsing? \\ Predictive Models for Error Analysis}

\author{Yang Janet Liu \and Tatsuya Aoyama \and Amir Zeldes \\
        Department of Linguistics \\ Georgetown University \\ 
        {\tt \{yl879, ta571, amir.zeldes\}@georgetown.edu}}

\begin{document}
\maketitle
\begin{abstract}
Despite recent advances in Natural Language Processing (NLP), hierarchical discourse parsing in the framework of Rhetorical Structure Theory remains challenging, and our understanding of the reasons for this are as yet limited. In this paper, we examine and model some of the factors associated with parsing difficulties in previous work: the existence of implicit discourse relations, challenges in identifying long-distance relations, out-of-vocabulary items, and more. 
In order to assess the relative importance of these variables, we also release two annotated English test-sets with explicit correct and distracting discourse markers associated with gold standard RST relations. Our results show that as in shallow discourse parsing, the explicit/implicit distinction plays a role, but that long-distance dependencies are the main challenge, while lack of lexical overlap is less of a problem, at least for in-domain parsing. 
Our final model is able to predict where errors will occur with an accuracy of $76.3$\% for the bottom-up parser and $76.6$\% for the top-down parser.
\end{abstract}

\section{Introduction}\label{sec:intro}

Powered by pretrained language models, recent advancements in NLP have led to rising scores on a myriad of language understanding tasks, especially at the sentence level. 
However, at the discourse level, where analyses require reasoning over multiple sentences, progress has been slower, with generalization to unseen domains remaining a persistent problem for tasks such as coreference resolution \cite{zhu-etal-2021-ontogum} and entity linking \cite{lin-zeldes-2021-wikigum}. 

One task which remains particularly challenging is hierarchical discourse parsing, which aims to reveal the structure of documents (e.g.~where parts begin and end, which parts are more important than others) and make explicit the relationship between clauses, sentences, and larger parts of the text, by labeling them as expressing a type of e.g.~\textsc{Causal}, \textsc{Elaboration}, etc. More specifically, hierarchical discourse parses identify  connections between elementary discourse units (EDUs, usually equated with propositions) in a text or conversation, classify their functions using a closed tag set, and form a recursive tree structure, which indicates the locally most prominent EDU in each tree or subtree. Figure 1 shows an example tree in the most popular hierarchical discourse formalism, Rhetorical Structure Theory (RST, \citealt{mann1988rhetorical}), in which the list of units $37$--$38$ is the most prominent (being pointed to by other units directly or indirectly), and discourse relation labels such as \textsc{cause} are identified using edge labels, whose definitions in RST are based on the rhetorical effect which the writer (or speaker) is thought to be conveying to the reader (or hearer).

\begin{figure}[t]
    \centering
    \includegraphics[width=\columnwidth]{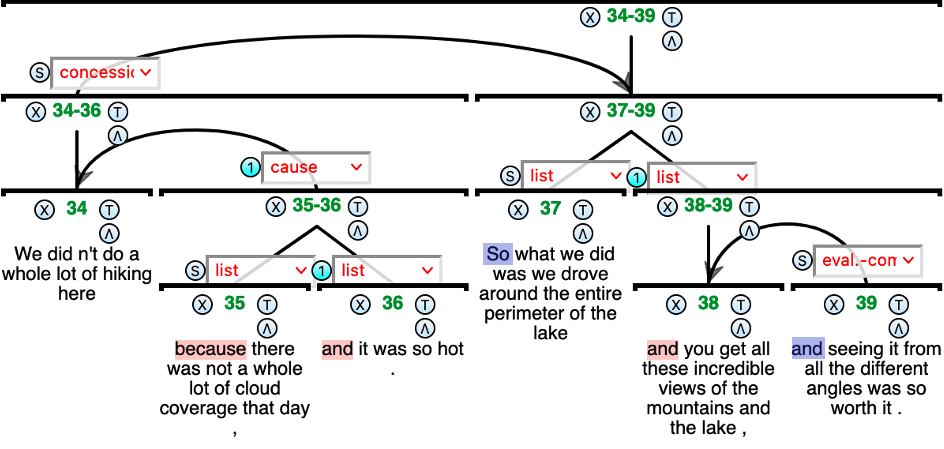}
    \caption{An RST analysis of a \textit{vlog} excerpt. Tokens highlighted in red are discourse markers associated with relations in the tree, while tokens highlighted in blue are distractors, with no corresponding relation. }
    \label{fig:rst-example}
    \vspace{-12pt}
\end{figure}

There is by now substantial evidence showing that even for a high resource language like English, state-of-the-art (SOTA) neural RST discourse parsers, whether employing a top-down or a bottom-up architecture, do not perform well across domains \cite{atwell-etal-2021-discourse,atwell-etal-2022-change,yu-etal-2022-rst,aoyama-etal-2023-gentle}, with some crucial tasks, such as predicting the most prominent Central Discourse Unit (CDU) of each document, performing at just 50\% \cite{liu-zeldes-2023-eacl}.
At the same time, we do not have a good understanding of what exactly prevents good performance---is it the fact that some relations are \textbf{well-marked} (for example, most \textsc{Contingency} relations are marked by the discourse marker (DM) \textit{if}, but most \textsc{Evaluation} relations lack a common marker)? Conversely, is the \textbf{presence of distracting markers} not associated with the correct relation (e.g.~an additional temporal marker such as \textit{then} inside a unit with a non-temporal function)? Alternatively, is it the difficulty in identifying high-level relations, between groups of multiple sentences or paragraphs, compared to less tricky intra-sentential relations between clauses? Or is it just the prevalence of out-of-vocabulary (OOV) items in test data?

In this paper, we would like to systematically evaluate the role of these and other factors contributing to errors in English RST discourse parsing. Our contributions include: 

\vspace{-2pt}
\begin{itemize}
\itemsep0.1em 
    \item Annotation and evaluation of the \texttt{dev}/\texttt{test} sets of the English RST-DT \cite{CarlsonEtAl2003} and GUM datasets \cite{Zeldes2017}, 
    for explicit relation markers, as well as distracting markers not signaling the correct relation;
    \item Parsing experiments with two different SOTA architectures to examine where degradation happens;
    \item Development and analysis of multifactorial models predicting where errors will occur and ranking importance for different variables;
    \item Qualitative and quantitative error analysis.
\end{itemize}
\vspace{-2pt}

\noindent Our results reveal that while explicit markers and distractors do play a role, the most significant predictor of difficulty is inter-sentential status and the specific relation involved. At the same time, our error analysis indicates that distractors often correspond to true discourse relations which are not included in the gold-standard tree, but may be included in alternative trees produced by other annotators. In addition, we find that OOV rate plays only a minor role, that architecture choice is presently not very important, and that genre continues to matter even when all other factors are known. All code and data are available at \url{https://github.com/janetlauyeung/NLPErrors4RST}.

\section{Related Work}
\label{sec:related-work}

\subsection{Discourse Structure in Discourse Parsing}


Discourse parsing is the task of identifying the coherence relations that hold between different parts of a text. Regardless of discourse frameworks or formalisms, identifying intra-sentential, inter-sentential, or inter-paragraph discourse relations may pose different levels of difficulty to parsers due to their various characteristics and levels of explicitness (e.g.~\citealt{zhao-webber-2021-revisiting,dai-huang-2018-improving,muller-etal-2012-constrained}). Intuitively, this becomes increasingly important for discourse parsing in a hierarchical framework such as RST, where long-distance relations are more frequent. 

Researchers have therefore been considering ways of dealing with long-distance relations for nearly twenty years, starting with the structure-informed model proposed by \citet{sporleder-lascarides-2004-combining} to tackle local and global discourse structures such as paragraphs. Other multi-stage parsing models, for example, as developed by \citet{joty-etal-2013-combining,joty-etal-2015-codra}, have taken into account the distribution and associated features of intra-sentential and inter-sentential relations, achieving competitive results for English document-level parsing. 

Later models expanded on these approaches by incorporating paragraph information to better capture high-level document structures. For instance, \citet{liu-lapata-2017-learning} proposed a neural model leveraging global context, enabling it to capture long-distance dependencies and achieving SOTA performance. \citet{yu-etal-2018-transition} used implicit syntactic features in a hierarchical RNN architecture. Active research continues on developing multi-stage parsing algorithms aiming at capitalizing on structural information at the sentence or paragraph-levels \cite{wang-etal-2017-two,lin-etal-2019-unified,Kobayashi2020TopDownRP,nishida-nakayama-2020-unsupervised,nguyen-etal-2021-rst}.

\subsection{Explicit and Implicit Relations in RST}

Unlike in hierarchical RST parsing, work on shallow discourse parsing in the framework of the Penn Discourse Treebank (PDTB, \citealt{prasad-etal-2014-reflections}), in which relations apply between spans of text without forming a tree, has long distinguished explicitly and implicitly marked discourse relations. Explicit relations are signaled by connectives such as `but' or `on the other hand', while implicit ones lack such marking. 
It is well-established that shallow parsing of explicit discourse relations is substantially easier due to the availability of connective signals, which, although not unambiguous, narrow down likely senses for relations. 
For example, the best systems from \citet{knaebel-2021-discopy} achieved an F1 score of $62.75$ on explicit relations and an F1 score of $40.71$ on implicit relations for Section $23$ of WSJ using PDTB v2 \cite{prasad-etal-2008-penn}. The DISRPT shared task created a relation classification task in 2021 \cite{zeldes-etal-2021-disrpt}, and the 2023 edition \cite{braud-etal-2023-disrpt} reported separate mean accuracy scores for explicit ($79.32$) and implicit ($50.85$) relations across six datasets in $4$ languages. 

RST datasets used in hierarchical discourse parsing do not make such a distinction, in part because RST trees include very high-level relations between entire sections of documents, which are less likely to be marked by such items. As a result, such a distinction is not available, meaning that we are in the dark regarding the prevalence and importance of such markers for RST parsing. 

We are aware of two prior works analyzing connectives for RST data: the RST Signalling Corpus (RST-SC, \citealt{DasTaboadaMcFetridge2019}) analyzes each relation in the English RST-DT dataset, indicating which relations were signaled by a DM (DMs roughly include the same items as PDTB connectives; see \citet{PDTB3-Annotation-Manual} and \citet{das2014rstsc-manual} for  complete inventories of markers). 
However, the data is limited to newswire material and does not provide an alignment of analyses to actual tokens, limiting the possibilities for model building (i.e.~we only know whether a DM was present somewhere, but not which token in the text it was or in which exact EDU it appeared). It also does not indicate whether DMs were present which \textit{\textbf{did not}} signal the relation in the tree (i.e.~distractors). Although previous efforts targeted DM tokens in RST-DT \cite{liu2019discourse} as well as such DM tokens in non-newswire texts \cite{liu-2019-beyond}, no previous study has examined the role of DMs in RST parsing. 

\citet{StedeNeumann2014} enriched an RST corpus of German with token-aligned connectives and the relations they signal, allowing investigation of their positions and the presence of distracting connectives. However, the annotations were not mapped to the RST relations in the corpus, making exact inferences again tricky, and the size of the corpus (32K tokens) precludes training high quality models. This corpus too is limited to the newspaper domain, which also motivates us to annotate genre-rich data, described in the next section.

Finally we note that data in other frameworks, including not only PDTB but also SDRT (Segmented Discourse Representation Theory, \citealt{AsherLascarides2003}), contains multiple concurrent discourse relations, providing information about the presence of competing or distracting relations. However, SDRT data does not include connective annotations, and apart from the coverage of RST-SC's overlapping data with the Wall Street Journal (WSJ) in PDTB, there is no way to extract a mapping between connectives and RST relations in any existing dataset (for attempts at aligning PDTB and RST-DT, see \citealt{DembergEtAl2019}).

In this paper, we therefore begin by creating hand-annotated data (using rstWeb, \citealt{gessler-etal-2019-discourse}) associating exact DM tokens with RST-style relations, or indicating their status as distractors, not associated with any relation in the gold tree. These latter DMs are especially interesting, since they could indicate that some parser errors are not exactly errors, instead corresponding to concurrent relations not present in the gold trees.

\section{Data}\label{sec:data}

To examine the role of explicit vs.~implicit relations in parsing errors, we first need to know which relations were explicitly signaled. To that end, we use PDTB's methodology to define explicit connectives. Note that RST papers often use the term DM without clear inventories; from this point on we will use `DM' for brevity, but strictly adhere to the PDTB English inventory. Specifically, we annotate data from the two largest RST corpora for English, covering the \texttt{test} set of RST-DT\footnote{RST-DT has no established separate \texttt{dev} set.} \cite{CarlsonEtAl2003} and the \texttt{test} and \texttt{dev} sets of GUM \cite{Zeldes2017} , with 1) \textbf{discourse markers} (including `distractor' DMs) and 2) \textbf{associated relations}, thereby attaching DMs to each relation they signal, or no relation. Table \ref{tab:rst-corpora-stats} gives an overview of the data.

\begin{table}[ht]
\small
\centering
\resizebox{7cm}{!}{%
\begin{tabular}{@{}l|cc@{}}
\toprule
& \textbf{RST-DT}        & \textbf{GUM v9}                    \\ \midrule
\textbf{\# of docs}             & $385$                    & $213$                                \\
\quad\textit{train/dev/test}         & \textit{$347$ / -- / $38$} & \textit{$165$ / $24$ / $24$}             \\
\textbf{\# of toks}             & $203,352$                & $203,780$                           \\
\textbf{\# of EDUs}             & $21,789$                 & $26,310$                             \\
\textbf{\# of genres}           & $1$                      & $12$                                 \\
\textbf{\# of relation labels}  & $78$                     & $32$                                 \\
\textbf{\# of relation classes} & $17$                     & $15$                                 \\ 
\textbf{\# of relation instances} & $18,630$               & $23,451$                             \\ \bottomrule
\end{tabular}%
}
\caption{Overview of the Largest English RST Corpora. }
\label{tab:rst-corpora-stats}
\vspace{-12pt}
\end{table}

\paragraph{Inter-Annotator Agreement}
To assess the reliability and quality of the human annotations, we conduct an inter-annotator agreement study on the \texttt{test} set of RST-DT and report average mutual F1 scores. The use of RST-DT can also facilitate some comparisons between the PDTB and RST frameworks as a number of documents from the WSJ section of the Penn Treebank \cite{MarcusSantoriniMarcinkiewicz1993} were annotated in both PDTB v3 and RST-DT.  
In total, we double-annotated $38$ documents, divided to overlap among three annotators. For DMs, the average F1 score was $95.2$, and for associated relations, the average F1 score given a DM was $96.7$. These scores indicate a high agreement between annotators for both tasks.

\paragraph{Automatic Parses}
In order to examine parsing errors from different architectures, we select two SOTA-performing parsers to obtain automatic parses: a \textsc{bottom-up} one from \citet{guz-carenini-2020-coreference}, using their best \texttt{SpanBERT-NoCoref} setting, and a \textsc{top-down} one from \citet{liu-etal-2021-dmrst} using XLM-RoBERTa-base \cite{conneau-etal-2020-unsupervised}. Following recommendations by \citet{morey-etal-2017-much}, we use the more stringent original Parseval metric on binary trees. Table \ref{tab:parser-results} shows reproduced 5-run average scores on both \texttt{test} sets.\footnote{Validation performance of each parser on both corpora is provided in Appendix \ref{appendix:training-and-val-performance}.} 
It is clear that scores of both architectures are neck and neck, which raises questions on whether, beyond numeric scores, they find similar or different data difficult.

\begin{table}[ht]
\resizebox{7cm}{!}{%
\begin{tabular}{@{}l|ccc|ccc@{}}
\toprule
\textit{corpora}                                        & \multicolumn{3}{c|}{\textbf{GUM v9}} & \multicolumn{3}{c}{\textbf{RST-DT}}  \\ \midrule
\textit{metrics}                                        & \textbf{S} & \textbf{N} & \textbf{R} & \textbf{S} & \textbf{N} & \textbf{R} \\ \midrule
\begin{tabular}[c]{@{}l@{}}\textsc{bottom-up}\\ \citet{guz-carenini-2020-coreference}\end{tabular} & $70.4$       & $57.7$       & $49.9$       & $76.5$       & $65.9$       & $54.8$       \\ \midrule
\begin{tabular}[c]{@{}l@{}}\textsc{top-down}\\ \citet{liu-etal-2021-dmrst}\end{tabular}  & $71.9$       & $58.9$       & $51.7$       & $76.5$       & $65.8$       & $54.8$       \\ \bottomrule
\end{tabular}%
}
\caption{Parsing Performance on GUM v9 and RST-DT \texttt{test} with Gold EDU Segmentation ($5$ run average). \textbf{S}=\textbf{S}pan (whether subtrees span the right EDUs); \textbf{N}=\textbf{N}uclearity (whether edges point the right way); \textbf{R}=\textbf{R}elation (whether labels are correct).}
\label{tab:parser-results}
\end{table}

\section{Analysis}

Strictly speaking, the types of errors that top-down and bottom-up parsers make are not identical: while bottom-up, and in particular shift-reduce parsers see analyzed preceding discourse units, grouped in a stack, and remaining discourse units in an upcoming queue, top-down parsers analyze a domain of ungrouped tokens to be split and determine the optimal split point and label for each decision. Because we want to analyze what promotes errors both across and for each architecture, we adopt an output-centric view, analyzing EDUs at which parsers do and do not make errors based on their properties in the completed gold vs.~predicted tree. 
At the same time, we do not want our results to be swayed by coincidental variations in neural models, which can have far-reaching consequences due to cascading errors. Instead, we train five models in each architecture, i.e.~five training runs, each with a different random seed producing a different initialization for the parser: if only one model fails to predict a relation, it may not be very hard, while $4$--$5$ errors would be indicative of genuinely hard relations.

\begin{figure}[t]
    \centering
    \includegraphics[scale=1,width=\columnwidth]{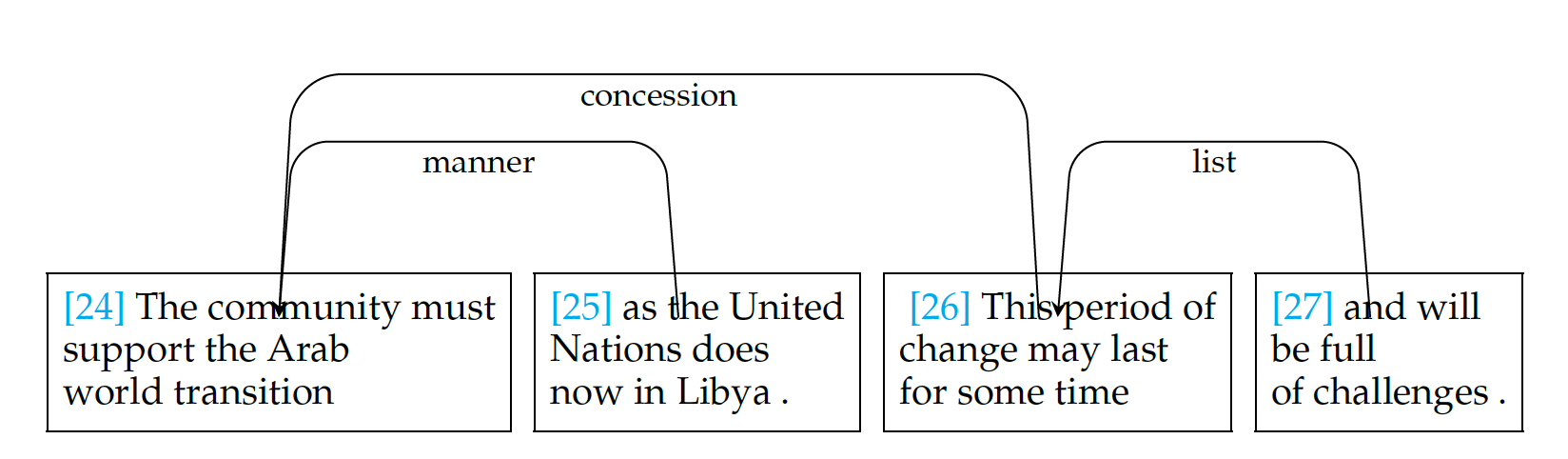}
    \caption{An Example of an RST Constituent Fragment converted into the Discourse Dependency Structure following \citet{li-etal-2014-text}.}
    \vspace{-8pt}
    \label{fig:gum-rst-dependencies-example}
\end{figure}

Additionally, since models ultimately confront different inputs as a result of such cascaded decisions, we  will use a dependency representation of both the gold and predicted RST trees, following the dependency conversion as defined by \citet{li-etal-2014-text},\footnote{The conversion code is available at \url{https://github.com/amir-zeldes/rst2dep}.} as exemplified in Figure \ref{fig:gum-rst-dependencies-example}. Although RST uses constituent discourse trees, focusing on each EDU and its dependencies will make it possible to make meaningful comparisons across models, and to intuitively understand how challenging EDUs are at any point in each document, regardless of whether or not they head large constituent structures. In Section \ref{subsec:other-factors} we will also incorporate the spanned domain of each head EDU's constituent block as an additional feature to assess the role of block size in predicting errors. 

\subsection{Explicit vs.~Implict Relations}\label{sec:main-factors}
\begin{table}[ht]
\centering
\resizebox{7cm}{!}{%
\begin{tabular}{@{}lrcccrc@{}}
\toprule
\multicolumn{1}{l|}{\textbf{}} &
  \multicolumn{1}{l}{\textbf{\begin{tabular}[c]{@{}l@{}}\# of\\ explicit\end{tabular}}} &
  \multicolumn{1}{l}{\textbf{\begin{tabular}[c]{@{}l@{}}explicit \\ prop.\end{tabular}}} &
  \multicolumn{1}{l}{\textbf{\begin{tabular}[c]{@{}l@{}}\# of\\ implicit\end{tabular}}} &
  \multicolumn{1}{l}{\textbf{\begin{tabular}[c]{@{}l@{}}implicit \\ prop.\end{tabular}}} &
  \multicolumn{1}{l}{\textbf{\begin{tabular}[c]{@{}l@{}}\# of \\ distractor\end{tabular}}} &
  \multicolumn{1}{l}{\textbf{\begin{tabular}[c]{@{}l@{}}distractor \\ prop.\end{tabular}}} \\ \midrule
\multicolumn{1}{l|}{\textbf{RST-DT}}       & $398$  & $17.0$\% & $1948$ & $83.0$\% & $81$  & $3.5$\%  \\ 
\multicolumn{1}{l|}{\textbf{GUM v9}}       & $1198$ & $21.7$\% & $4332$ & $78.3$\% & $174$ & $3.1$\%  \\ \midrule 
\multicolumn{1}{l|}{\textit{academic}}     & $73$   & $16.1$\% & $380$  & $83.9$\% & $13$   & $2.9$\%                     \\
\multicolumn{1}{l|}{\textit{bio}}          & $66$   & $18.4$\% & $292$  & $81.6$\% & $11$   & $3.1$\%                     \\
\multicolumn{1}{l|}{\textit{conversation}} & $100$  & $12.9$\% & $674$  & \textcolor{blue}{\textbf{$87.1$\%}} & $23$   & $3.0$\%                     \\
\multicolumn{1}{l|}{\textit{fiction}}      & $116$  & $23.7$\% & $374$  & $76.3$\% & $15$   & $3.1$\%                     \\
\multicolumn{1}{l|}{\textit{interview}}    & $80$   & $20.2$\% & $317$  & $79.8$\% & $8$    & $2.0$\%                     \\
\multicolumn{1}{l|}{\textit{news}}         & $73$   & $18.1$\% & $331$  & $81.9$\% & $7$    & $1.7$\%                     \\
\multicolumn{1}{l|}{\textit{reddit}}       & $147$  & $28.3$\% & $373$  & $71.7$\% & $20$   & $3.8$\%                     \\
\multicolumn{1}{l|}{\textit{speech}}       & $84$   & $19.1$\% & $356$  & $80.9$\% & $9$    & $2.0$\%                     \\
\multicolumn{1}{l|}{\textit{textbook}}     & $95$   & $21.3$\% & $352$  & $78.7$\% & $9$    & $2.0$\%                     \\
\multicolumn{1}{l|}{\textit{vlog}}         & $180$  & \textcolor{blue}{\textbf{$35.8$\%}} & $323$  & $64.2$\% & $38$   & \textcolor{blue}{\textbf{$7.6$\%}}                     \\
\multicolumn{1}{l|}{\textit{voyage}}       & $69$   & $22.4$\% & $239$  & $77.6$\% & $9$    & $2.9$\%                     \\
\multicolumn{1}{l|}{\textit{whow}}         & $115$  & $26.4$\% & $321$  & $73.6$\% & $12$   & $2.8$\%                     \\ \midrule
\multicolumn{1}{l|}{mean}                  & $99.8$ & $21.9$\% & $361$  & $78.1$\% & $14.5$ & $3.1$\%                     \\ \bottomrule
\end{tabular}%
}
\caption{Distribution of Explicit and Implicit Relations as well as EDUs with Distracting DMs in RST-DT \texttt{test} and \texttt{dev}+\texttt{test} of GUM v9. }
\vspace{-5pt}
\label{tab:exp-vs-imp-vs-distracting-stats}
\end{table}


Table \ref{tab:exp-vs-imp-vs-distracting-stats} shows the distribution of explicit or unmarked relations across the genres in the \texttt{dev}+\texttt{test} sets of GUM v9 and in comparison to RST-DT's \texttt{test} set, for each relation class and overall. The results for RST-DT are consistent with previous work, with $17.0$\% of test data relations being marked, similarly to the $18.2$\% identified by \citet{DasTaboada2017} for the entire corpus (but not anchored to specific tokens). An examination of distributions by genre in GUM reveals some differences, highlighted in Table \ref{tab:exp-vs-imp-vs-distracting-stats}, with \textit{\textbf{vlog}} exhibiting the most explicit relations, and \textit{\textbf{conversation}} the fewest, raising the possibility that it may be more challenging for parsers. And in fact, \citet{liu-zeldes-2023-eacl} pointed to  \textit{conversation} as the worst-performing genre at all metric levels using an older version of the corpus (v8), which had less  \textit{conversation} data compared to GUM v9. 

Looking at the presence of `distractor' connectives, which are not associated with one of the gold relations in the tree, we see that \textit{\textbf{vlog}} is the most prone to such cases, again raising the question of whether these may pose a problem for parsers, which may identify \textbf{a possibly correct relation that is not prioritized by the gold tree}. This situation appears to be infrequent in the WSJ data from RST-DT, which has only $81$ such cases ($3.5$\%). Taking a closer look at the types of distractors across genres in GUM, we see that the most frequent types are `and', `but', and `so', which are highly ambiguous and common in conversational data such as \textit{vlog} and \textit{conversation}. 

Regarding the most and least explicitly signaled relation classes in GUM v9, Table \ref{tab:gum-rel-class-stats} reveals that \textsc{Contingency} is the most explicitly marked class due to the use of the DM `if', and that the least explicitly signaled classes are \textsc{Attribution} and \textsc{Organization}. The former is almost always signaled by speech verbs (a verb such as `say' or `argue') and the latter mostly by document layout and graphical features in written texts, or by back-channeling in conversation data.
It is also worth noting that instances of \textsc{Evaluation}, \textsc{Restatement}, and \textsc{Topic} (used predominantly for question-answer pairs) are mostly \textit{\textbf{not}} signaled by a discourse marker.

\begin{table}[ht]
\centering
\resizebox{6.5cm}{!}{%
\begin{tabular}{@{}l|rrrr@{}}
\toprule
\textbf{\begin{tabular}[c]{@{}l@{}}relation \\ class\end{tabular}} &
  \multicolumn{1}{l}{\textbf{\begin{tabular}[c]{@{}l@{}}\# of\\ explicit\end{tabular}}} &
  \multicolumn{1}{l}{\textbf{\begin{tabular}[c]{@{}l@{}}explicit \\ prop.\end{tabular}}} &
  \multicolumn{1}{l}{\textbf{\begin{tabular}[c]{@{}l@{}}\# of \\ implicit\end{tabular}}} &
  \multicolumn{1}{l}{\textbf{\begin{tabular}[c]{@{}l@{}}implicit \\ prop.\end{tabular}}} \\ \midrule
\textsc{ROOT}         & $0$   & $0.0$\%  & $48$   & $100.0$\% \\
\textsc{Adversative}  & $222$ & $55.5$\% & $178$  & $44.5$\%  \\
\textsc{Attribution}  & $0$   & $0.0$\%  & $292$  & \textcolor{blue}{\textbf{$100.0$\%}} \\
\textsc{Causal}       & $131$ & $53.5$\% & $114$  & $46.5$\%  \\
\textsc{Context}      & $143$ & $31.8$\% & $306$  & $68.2$\%  \\
\textsc{Contingency}  & $99$  & \textcolor{blue}{\textbf{$91.7$\%}} & $9$    & $8.3$\%   \\
\textsc{Elaboration}  & $64$  & $5.8$\%  & $1049$ & $94.2$\%  \\
\textsc{Evaluation}   & $4$   & $1.7$\%  & $231$  & $98.3$\%  \\
\textsc{Explanation}  & $44$  & $12.5$\% & $308$  & $87.5$\%  \\
\textsc{Joint}        & $409$ & $37.2$\% & $689$  & $62.8$\%  \\
\textsc{Mode}         & $52$  & $45.2$\% & $63$   & $54.8$\%  \\
\textsc{Organization} & $0$   & $0.0$\%  & $331$  & \textcolor{blue}{\textbf{100.0\%}} \\
\textsc{Purpose}      & $21$  & $10.7$\% & $176$  & $89.3$\%  \\
\textsc{Restatement}  & $6$   & $3.8$\%  & $150$  & $96.2$\%  \\
\textsc{Same-Unit}    & $1$   & $0.3$\%  & $289$  & $99.7$\%  \\
\textsc{Topic}        & $2$   & $2.0$\%  & $99$   & $98.0$\%  \\ \bottomrule
\end{tabular}%
}
\caption{Distribution of Explicit and Implicit Relations across Relation Classes in \texttt{dev}+\texttt{test} of GUM v9. }
\label{tab:gum-rel-class-stats}
\vspace{-5pt}
\end{table}

\begin{figure}[ht]
    \centering
    \includegraphics[width=\columnwidth]{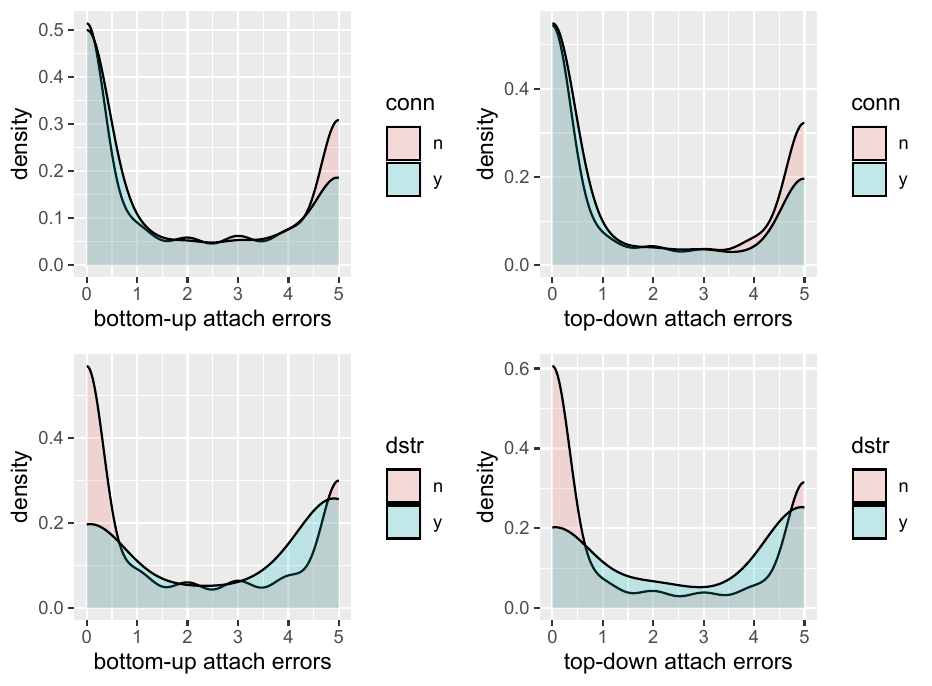} 
    \caption{Attachment Error Count Density with and without DMs or Distractors for Each Architecture in \texttt{dev+test} of GUM v9. }
    \label{fig:densities}
\end{figure}

\begin{figure}[ht]
    \centering
    \includegraphics[width=\columnwidth]{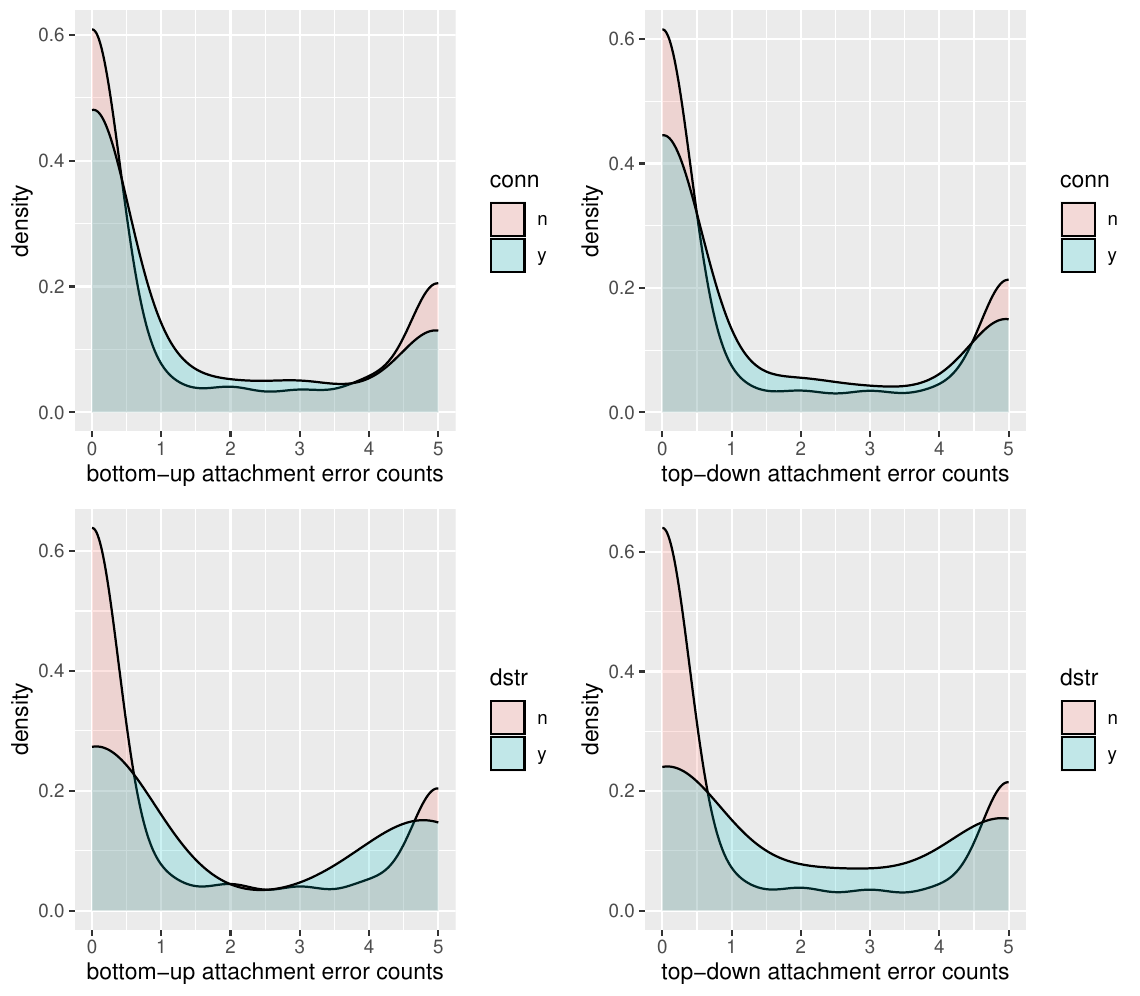} 
    \caption{Attachment Error Count Density with and without DMs or Distractors for Each Architecture in \texttt{test} of RST-DT. }
    \vspace{-5pt}
    \label{fig:densities-RSTDT}
\end{figure}

With these descriptive statistics in hand, we can examine each parser's performance on explicit/implicit relations, as well as on EDUs with a distracting DM in either the source or target of the relation (we must consider both ends, since many DMs can mark either a source or target such as `but' and `so'). Figure \ref{fig:densities} shows the density of relations incurring between $0$ and $5$ attachment errors (disregarding labels) in each architecture for GUM, 
broken down by whether a DM marks the relation (top) and whether a distracting DM is present (bottom). The figure reveals several important facts: firstly, DMs are unsurprisingly associated with fewer errors ($t$=$-7.29$, $D$=$0.23$, $p$<$0.0001$), with lack of connectives affecting top-down models slightly more severely ($\chi^2$=$3.95$, $\phi$=$0.14$, $p$<$0.05$). Secondly, lack of distractors is associated with having fewer errors ($t$=$5.0718$, $D$=$0.37$, $p$<$0.0001$), and this is more pronounced for the bottom-up architecture, but the difference between architectures is not significant here.\footnote{That said, we recognize that there are also more differences between these parsers than just the top-down/bottom-up distinction, so it is possible that with a broader sample of parsers, more differences would emerge.} Figure \ref{fig:densities-RSTDT} shows the same kind of density plots for RST-DT.

Although it seems obvious that explicitness will facilitate parsing and that distractors should be harmful, it is an open question whether such markers will remain important once we know about other factors known to cause problems, such as OOV items, EDU text length, and intra-sentential status. To compare these, we construct several regression models predicting the number of errors. 
Because the distribution of error numbers is U-shaped (many cases with zero or five errors, few in the middle), as shown in Figures \ref{fig:densities}--\ref{fig:densities-RSTDT}, we cannot use traditional gaussian models, which assume a roughly normal distribution of the data. Instead, we use mixed effects Beta regression, which is suited to U-shaped data, with a random effect for document identity, and re-scale the number of \textbf{attachment or relation errors} to the range $0$--$1$, where $1$ means the max $5$ model errors. Table \ref{tab:regression-model-results} shows significance for each predictor in each model.\footnote{Significance for \texttt{genre}, a multi-nominal feature, is computed via a likelihood ratio test comparing the model with and without this predictor.}

\begin{table*}[tbh]
\centering
\resizebox{\textwidth}{!}{%
\begin{tabular}{r|cc|cc|cc|cc|cc|cc}
\toprule
corpus & \multicolumn{6}{|c|}{\textbf{GUM v9}} & \multicolumn{6}{|c}{\textbf{RST-DT}} \\
  \midrule
 architecture & bot-up & top-down & bot-up & top-down & bot-up & top-down & bot-up & top-down & bot-up & top-down & bot-up & top-down \\ 
  \midrule
\texttt{dm} & $<$.$001$*** & $<$.$001$*** & $0.059$ & $0.074$ & $0.003$** & $0.005$** & $0.988$ & $0.002$** & $0.244$ & $<$.$001$*** & $0.445$ & $<$.$001$*** \\ 
  \texttt{distractor} & $<$.$001$*** & $<$.$001$*** & $<$.$001$*** & $<$.$001$*** & $<$.$001$*** & $<$.$001$*** & $<$.$001$*** & $<$.$001$*** & $<$.$001$*** & $<$.$001$*** & $<$.$001$*** & $<$.$001$*** \\ 
  \texttt{subord} &  &  & $<$.$001$*** & $<$.$001$*** & $<$.$001$*** & $<$.$001$*** &  &  & $<$.$001$*** & $<$.$001$*** & $<$.$001$*** & $<$.$001$*** \\ 
  \texttt{length} &  &  &  &  & $<$.$001$*** & $<$.$001$*** &  &  &  &  & $<$.$001$*** & $<$.$001$*** \\ 
  \texttt{oov} &  &  &  &  & $0.115$ & $0.262$ &  &  &  &  & $0.944$ & $0.563$ \\ 
  \texttt{genre} &  &  &  &  & $<$.$001$*** & $<$.$001$*** &  &  &  &  &  &  \\ 
   \bottomrule
\end{tabular}
}
\caption{Results of the Regression Models for GUM v9 and RST-DT from both Architectures.}
\label{tab:regression-model-results}
\end{table*}

Looking first at GUM on the left, Table \ref{tab:regression-model-results} shows that, when given only DMs and distractors, both features are significant in predicting errors above a per-document random effect baseline, for both architectures. In other words, predicting implicit relations is unsurprisingly harder in RST, just as it is for PDTB-style shallow discourse parsing, and distractors make things even harder.

However, adding the subordination feature (the second and third pairs of models from the left for GUM v9), which indicates whether an EDU is in a subordinate clause (and therefore likely to have an intra-sentential relation), removes the significance of the presence of a DM (but not of distractors). This suggests DMs are less important in predicting errors (or lack thereof) than intra-sentential status. Adding some more predictors, a fuller model with EDU length, OOV rate (the percentage of lexical items not seen during training per EDU), and genre does not remove the significance of subordination status, and shows that OOV rate is not a significant predictor in this setting. The more complex models with 6 features also restore some significance for DMs, albeit to a lesser degree than other predictors.

Moving to RST-DT, we see a similar pattern, except for a surprising difference between architectures: in the mixed effects model, presence of a DM is \textbf{\textit{not}} a significant predictor for the bottom-up architecture, while it is significant for top-down. This pattern is repeated across all sets of features on the right side of Table \ref{tab:regression-model-results}. 
For RST-DT, since we do not have gold syntactic dependency trees, we use gold intra-sentential relation status to represent the \texttt{subord} feature. This feature remains highly significant in all models across architectures. Finally, adding all the features to the right-most models (excluding \texttt{genre}, since RST-DT is all newswire), OOV rate again fails to reach significance, while all other features are significant, except for DMs for the bottom-up architecture models.

These numbers suggest several things: first and most important, while DMs may be somewhat important, some representation of intra-sentential status is the more robust predictor of parsing errors. This effect persists even if we know about other plausible features, such as EDU length and OOV rate. This observation fits with the line of work mentioned above on multi-stage models for RST parsing, which attempt to learn separate models for intra-sentential and inter-sentential or inter-paragraph models (e.g.~\citealt{Kobayashi2020TopDownRP}). Although joint models can perform well on all levels regardless, we can confirm that there are substantial differences between these types.

In terms of architecture differences, results for RST-DT suggest more sensitivity to DMs for top-down models, but this result is not reproduced in GUM. Finally, all models are sensitive to distractors, which raises questions about the nature of this sensitivity---what kinds of errors are parsers making, and more specifically are they predicting relations corresponding to distractor DMs? We address these questions in the next sections.

\subsection{Predicting Parsing Errors}\label{subsec:other-factors}

The results in the previous section quantify  the importance of different characteristics of discourse relations in promoting errors, and the relative difficulty of implicit relations in SOTA English RST parsing. 

However, the linear model comparing the significance of explicit DMs, distractors, and features such as EDU length or OOV rate is rather naive and leaves out a variety of potentially relevant properties of subtrees, such as total number of attached discourse units (which could contribute to ambiguity), or the gold relation to be predicted---some relations are easier to recognize or are less ambiguous, and some relations have high prior likelihood, making guessing them a safe bet. Although these properties may not be useful for realistic prediction of errors when we do not have a gold parse, they can be of interest for understanding tree properties which are difficult for parsers to get right.

To make matters even more complex, the factors mentioned above interact in subtle ways with each other and with explicit marking status. For example, \textsc{Contingency} relations are easy to recognize thanks to the reliable DM `if' as in \ref{ex:if}, but this is not always the case, as in \ref{ex:inversion} which uses subject-verb inversion to mark a conditional. Some relations are almost never marked by DMs, but may still be easy, such as \textsc{Attribution}, which can be identified via speech verbs, as in \ref{ex:attribution}. 

\ex. \small [Um \textcolor{blue}{\textbf{if}} you don't want to do a tour of Pittock Mansion,]$\xrightarrow[]{\text{gold:\textsc{Contingency}}}$[I'd still recommend like taking the trail up there]\textsubscript{GUM\_vlog\_portland} \label{ex:if}

\ex. \small [``\textcolor{blue}{\textbf{Had it happened}} an hour later]$\xrightarrow[]{\text{gold:\textsc{Contingency}}}$[It would have been much worse]\textsubscript{GUM\_news\_crane} \label{ex:inversion} 

\ex. \small [Any judge in this country would \textcolor{blue}{\textbf{agree}}]$\xrightarrow[]{\text{gold:\textsc{Attribution}}}$[that opening and closing statements along are not a trial.]\textsubscript{GUM\_speech\_impeachment} \label{ex:attribution}

This complexity means that a realistic model of difficult parsing environments may need to consider more variables, and the interactions mean that a simple linear model cannot capture the rich patterns in the data. In this section, we therefore use XGBoost \cite{XGBOOST}, a highly accurate ensemble gradient boosting framework which is able to harness arbitrary interactions between features and is highly regularized to prevent overfitting, meaning it can be expected to find a near-optimal mapping of our variables to parser error occurrences. For this experiment, we will attempt to predict `hard' EDUs, which we define as EDUs which most models predict incorrectly.

However, it is not immediately clear what kinds of features we should allow the model to use: on the one hand, we would like to know what constellations in gold RST trees are difficult, including the gold relation label or the relative importance of being a leaf node vs.~a hub with many dependents, as well as the contributions of DMs and distractors. On the other hand, in a realistic scenario we would not be able to know whether a DM is a distractor without knowing the gold relation, and we would not know how many dependents a node really has.

\begin{figure}[ht]
    \centering
    \includegraphics[width=\columnwidth]{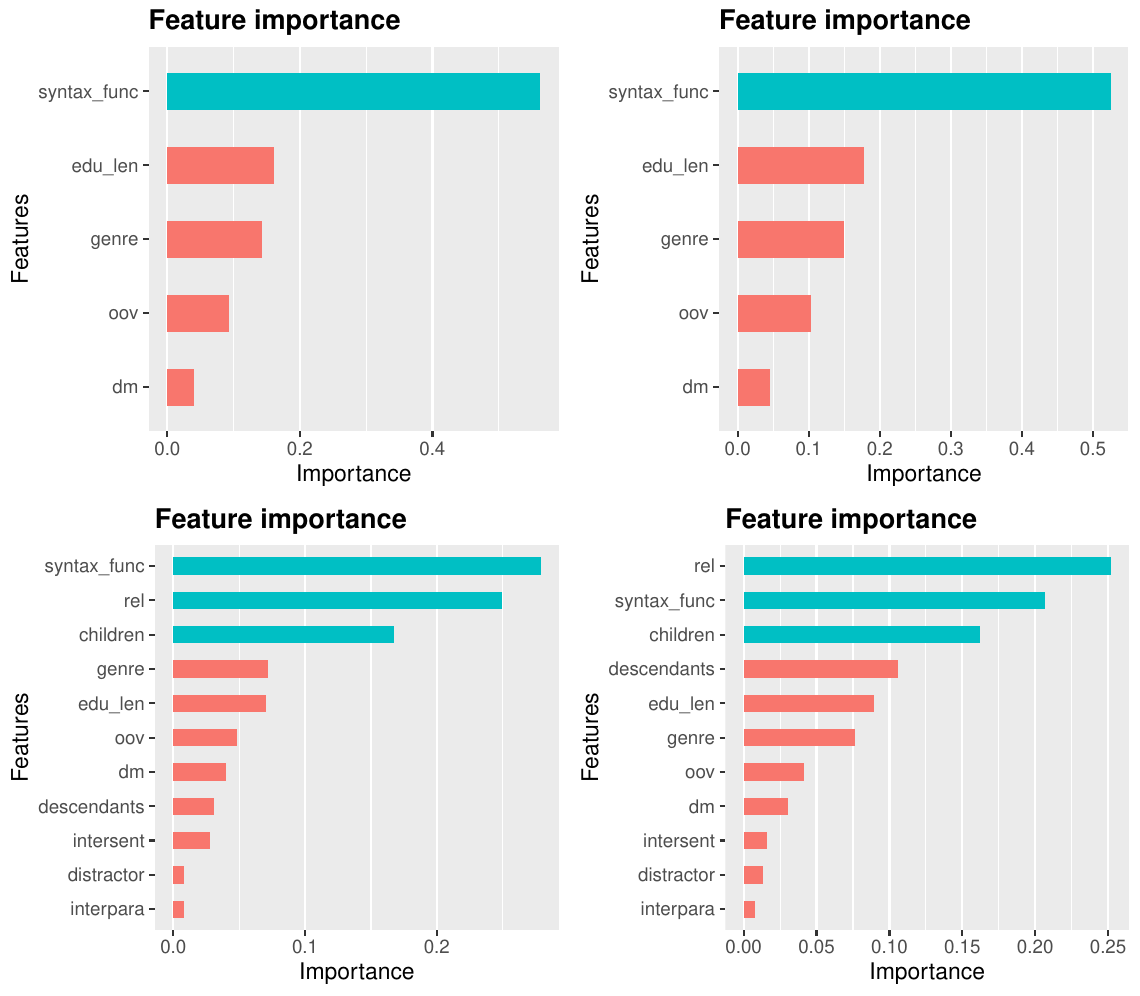} 
    \caption{Feature Importances for the \textsc{\textbf{Realistic}} (top) and \textsc{\textbf{Full}} (bottom) XGBoost Models for GUM from both \textsc{bottom-up} (left) and \textsc{top-down} (right) Architectures. Very important features are highlighted in \textcolor{teal}{\textbf{teal}}.}
    \label{fig:xgboost}
\end{figure}

We thus construct two models: the \textsc{\textbf{Realistic}} model only has access to features that can reasonably be predicted without the gold parse, including EDU length in tokens, presence of DMs (whether helpful or distracting), the incoming syntactic dependency relation (which can be predicted by a syntax parser), the OOV rate, and genre. The \textsc{\textbf{Full}} model, by contrast, has access to all gold features, including the gold relation class, intra-/inter-sentential status, DM vs.~distractor presence etc. The first model is more relevant for realistic scenarios in which we want to diagnose where parser errors are more likely (or how many we might incur), while the second is more helpful for understanding what is hard in an RST graph given the gold graph itself. Note that neither model is fed features from any outputs of the parser models above: the parsers are only used to compute the number of errors at each point, which the XGBoost model attempts to predict. Figure \ref{fig:xgboost} gives an analysis of feature importances using classification gain\footnote{Because XGBoost relies on gradient boosting with tree-based learners, the effect of variable interactions is computed within the classification gain metric, which is often used to estimate feature importance (see e.g.~\citealt{ShangEtAl2019}).} for both the \textsc{Realistic} and the \textsc{Full} models, which score $67.3$\% and $76.3$\% respectively over a majority baseline score of $58.3$\%, which predicts that RST parsers will never be wrong, for the bottom-up architecture. For top-down, the scores of the two models are $65.3$\% and $76.6$\% respectively.

The XGBoost library's plots automatically highlight the most important features for both parser architectures, which for the \textsc{\textbf{Realistic}} model is only \textbf{the syntactic function of the EDU}. This likely indicates the overwhelming importance of knowing whether an EDU has a typical intra-sentential role, such as a relative or adverbial clause, which is likely to be predicted correctly. The next features begin with length (short EDUs are likely to have similar ones attested in training data compared to long ones), then genre (since some genres are harder), and only then the typical NLP difficulty predictor, the OOV rate (which is slightly less useful when EDU length is also known, since the two correlate). The last feature, presence of DMs, is still useful but less so, especially since it folds in occurrences of helpful and distracting DMs. There are no substantial differences between top-down and bottom-up here for GUM v9. 

Turning to the \textsc{\textbf{Full}} model, we see that syntactic function is still very important: it beats gold label for bottom-up models and follows it for top-down. Some relations are easier than others, or different subsequent conditions apply to them, and this matters about as much as the syntactic attachment type. 
Number of children (a measure of tree centrality vs.~leaf status) is third, only then followed by length and genre, which are still quite helpful. Number of descendants (which is correlated with children) follows for top-down, but is far lower for bottom-up parsers. We then see OOV rate outranking DMs, which outrank less important features, such as the no longer crucial inter-sentential/inter-paragraph status, which are also highly correlated with some of the features above (syntax for the former, number of children for the latter, since many children are typical of paragraph head units). Finally distractors are second to last, far below DMs, also because they are rare.

These models indicate that predicting errors without knowing the gold tree is challenging, but a gain of $7$--$9$\% over baseline is still possible, mainly by looking at syntactic structure, which indicates inter-/intra-sentential status---a predictor much more valuable than DM marking. By contrast, when looking at gold trees, hard parts can most easily be associated with hard relations and syntactic environments, but combining all of the available features leads to an impressive ability to predict where parser models will likely go wrong, with $\sim$$18$\% gain over baseline.

\subsection{The Nature and Meaning of Distractors} 

Although the previous results suggest distractors play a minor role, their independent correlation with errors and the fact that DMs are generally relevant to discourse relations, raise questions regarding their very existence: why do they appear and how exactly do they affect parsers? 

To begin with the second question, we examined the $174$ distractors in GUM. For most bottom-up models, $108$/$174$ ($62.1$\%) were still erroneous, and $107$/$174$ ($62.1$\%) instances from the top-down models were erroneous. We then decided to manually label whether the majority model-predicted label was consistent with the distractor: if the gold relation is \textsc{Elaboration}, the distractor is \textit{but}, and the prediction is \textsc{Adversative}, then prediction is consistent with the distractor, but if the prediction is \textsc{Contingency}, then it is not. We use PDTB's mapping of connectives to classes to match DMs to relations. 

For $74$/$108$ cases ($68.5$\%) from the bottom-up models and $68$/$107$ cases ($63.6$\%) from the top-down models, the majority label was consistent with the distractor---in other words, the parser may be predicting based on a DM which would normally signal a competing relation. 
This brings us to the second question: if the relations signaled by distractors are incorrect, why are the distractors present? As an example, we consider two such cases from GUM, shown in \ref{ex:dist-if}--\ref{ex:dist-but}.

\ex. \small {[\textcolor{blue}{\textbf{if}} Steven didn't see it as weird]$\xrightarrow[\text{pred:\textsc{Contingency}}]{\text{gold:\textsc{Explanation}}}$[why should it bother us?]\textsubscript{GUM\_fiction\_teeth}} \label{ex:dist-if}

\ex. \small {[so the reason seems to be that there are things out there that put even these kaiju to shame ]$\xleftarrow[\text{pred:\textsc{Adversative}}]{\text{gold:\textsc{Evaluation}}}$[\textcolor{blue}{\textbf{But}} even this presents a problem ]\textsubscript{GUM\_reddit\_monsters} }\label{ex:dist-but}

In \ref{ex:dist-if}, the gold tree has the `if'-clause as a justification for why it `shouldn't bother us', which makes sense pragmatically; but formally, the clause seems like a legitimate conditional marked by \textit{if}, and parsers predict \textsc{Contingency}. In \ref{ex:dist-but}, the annotation focuses on the evaluative meaning of the words `a problem', while parsers, probably provoked by \textit{But}, predict \textsc{Adversative}.

We thus suspect that multiple, concurrent relations may actually hold in data where distractors appear, which is a standard possibility in frameworks like PDTB, where relations are identified based on the presence of DMs. If this applies in RST as well, then in a sense, such parser errors are not really errors at all. Because RST enforces a strict tree constraint, the only way to find out would be to look at alternative RST trees.

In order to do just this, we utilize RST-DT's official double-annotated subset, which has trees from a second annotator for $53$ documents. This subset overlaps only 5 documents in the RST-DT \texttt{test} set, which contain only $12$ distractors, meaning that the scope of this last analysis is limited; however, in examining these $12$ distractors, we discovered that $75$\% ($9$/$12$) actually corresponded to relations \textbf{selected as the primary RST relations by the second annotator} in the double annotated data. In other words, the double annotated data confirms that, at least in the case of the RST-DT \texttt{test} set, a large majority of distractors do in fact correspond to multiple concurrent relations, which were identified by an experienced RST annotator.

\section{Conclusion} 

This study has several important implications. Firstly and unsurprisingly, the explicit/implicit distinction from shallow discourse parsing is mirrored in RST parsing difficulty, and the dataset released in this paper can help study it further. However, explicit marking is clearly less consequential than intra-sentential status, with which explicitness it correlated. Secondly, OOV rate plays a less important role than we initially suspected, while genre effects remain robust, suggesting that diverse genres may matter more than subject matter. Our results also indicate that current architectures do not differ substantially in what they get right or wrong, and with scores being so similar, differences reduce to computational efficiency and personal preference.

Finally, the study of distractors suggest that RST's tree constraint may mix some cases of multiple concurrent relations with parsing errors, when parsers are actually identifying viable relations. This suggests that we may want to consider ways of allowing and adding concurrent relations to RST parses.

We also note that although the error prediction models evaluated in Section \ref{subsec:other-factors} were primarily developed in order to gain a greater understanding of the issues in discourse parsing, they could have some practical applications.\footnote{We thank an anonymous reviewer for noting this.} Predicting regions of low certainty in discourse parses can: 1) assist by highlighting low confidence regions in user-facing downstream applications; 2) flag potential problems during annotation of resources, especially when relying on NLP \cite{gessler-etal-2020-amalgum} or less trained annotators/crowd workers \cite{scholman-etal-2022-design,pyatkin2023design}; and 3) help guide additional resource acquisition, either automatically using active learning (to prioritize documents predicted to have parsing problems for manual annotation, cf.~\citealt{gessler-etal-2022-midas}) or using qualitative evaluation in deciding what data to collect in terms of the relative importance of genres, presence of OOV items, etc.

\bibliography{anthology,custom}

\begin{thebibliography}{50}
\expandafter\ifx\csname natexlab\endcsname\relax\def\natexlab#1{#1}\fi

\bibitem[{Aoyama et~al.(2023)Aoyama, Behzad, Gessler, Levine, Lin, Liu, Peng,
  Zhu, and Zeldes}]{aoyama-etal-2023-gentle}
Tatsuya Aoyama, Shabnam Behzad, Luke Gessler, Lauren Levine, Jessica Lin,
  Yang~Janet Liu, Siyao Peng, Yilun Zhu, and Amir Zeldes. 2023.
\newblock \href {https://aclanthology.org/2023.law-1.17} {{GENTLE}: A
  genre-diverse multilayer challenge set for {E}nglish {NLP} and linguistic
  evaluation}.
\newblock In \emph{Proceedings of the 17th Linguistic Annotation Workshop
  (LAW-XVII)}, pages 166--178, Toronto, Canada. Association for Computational
  Linguistics.

\bibitem[{Asher and Lascarides(2003)}]{AsherLascarides2003}
Nicholas Asher and Alex Lascarides. 2003.
\newblock \emph{Logics of Conversation}.
\newblock Studies in Natural Language Processing. Cambridge University Press,
  Cambridge.

\bibitem[{Atwell et~al.(2021)Atwell, Li, and
  Alikhani}]{atwell-etal-2021-discourse}
Katherine Atwell, Junyi~Jessy Li, and Malihe Alikhani. 2021.
\newblock \href {https://aclanthology.org/2021.sigdial-1.34} {Where are we in
  discourse relation recognition?}
\newblock In \emph{Proceedings of the 22nd Annual Meeting of the Special
  Interest Group on Discourse and Dialogue}, pages 314--325, Singapore and
  Online. Association for Computational Linguistics.

\bibitem[{Atwell et~al.(2022)Atwell, Sicilia, Hwang, and
  Alikhani}]{atwell-etal-2022-change}
Katherine Atwell, Anthony Sicilia, Seong~Jae Hwang, and Malihe Alikhani. 2022.
\newblock \href {https://doi.org/10.18653/v1/2022.findings-acl.68} {The change
  that matters in discourse parsing: Estimating the impact of domain shift on
  parser error}.
\newblock In \emph{Findings of the Association for Computational Linguistics:
  ACL 2022}, pages 824--845, Dublin, Ireland. Association for Computational
  Linguistics.

\bibitem[{Braud et~al.(2023)Braud, Liu, Metheniti, Muller, Rivi{\`e}re,
  Rutherford, and Zeldes}]{braud-etal-2023-disrpt}
Chlo{\'e} Braud, Yang~Janet Liu, Eleni Metheniti, Philippe Muller, Laura
  Rivi{\`e}re, Attapol Rutherford, and Amir Zeldes. 2023.
\newblock \href {https://aclanthology.org/2023.disrpt-1.1} {The {DISRPT} 2023
  shared task on elementary discourse unit segmentation, connective detection,
  and relation classification}.
\newblock In \emph{Proceedings of the 3rd Shared Task on Discourse Relation
  Parsing and Treebanking (DISRPT 2023)}, pages 1--21, Toronto, Canada. The
  Association for Computational Linguistics.

\bibitem[{Carlson et~al.(2003)Carlson, Marcu, and Okurowski}]{CarlsonEtAl2003}
Lynn Carlson, Daniel Marcu, and Mary~Ellen Okurowski. 2003.
\newblock Building a {D}iscourse-{T}agged {C}orpus in the {F}ramework of
  {R}hetorical {S}tructure {T}heory.
\newblock In \emph{Current and New Directions in Discourse and Dialogue}, Text,
  Speech and Language Technology 22, pages 85--112. Kluwer, Dordrecht.

\bibitem[{Chen and Guestrin(2016)}]{XGBOOST}
Tianqi Chen and Carlos Guestrin. 2016.
\newblock \href {https://doi.org/10.1145/2939672.2939785} {{XGBoost}: A
  scalable tree boosting system}.
\newblock In \emph{Proceedings of the 22nd ACM SIGKDD International Conference
  on Knowledge Discovery and Data Mining}, KDD '16, page 785–794, New York,
  NY, USA. Association for Computing Machinery.

\bibitem[{Conneau et~al.(2020)Conneau, Khandelwal, Goyal, Chaudhary, Wenzek,
  Guzm{\'a}n, Grave, Ott, Zettlemoyer, and
  Stoyanov}]{conneau-etal-2020-unsupervised}
Alexis Conneau, Kartikay Khandelwal, Naman Goyal, Vishrav Chaudhary, Guillaume
  Wenzek, Francisco Guzm{\'a}n, Edouard Grave, Myle Ott, Luke Zettlemoyer, and
  Veselin Stoyanov. 2020.
\newblock \href {https://doi.org/10.18653/v1/2020.acl-main.747} {Unsupervised
  cross-lingual representation learning at scale}.
\newblock In \emph{Proceedings of the 58th Annual Meeting of the Association
  for Computational Linguistics}, pages 8440--8451, Online. Association for
  Computational Linguistics.

\bibitem[{Dai and Huang(2018)}]{dai-huang-2018-improving}
Zeyu Dai and Ruihong Huang. 2018.
\newblock \href {https://doi.org/10.18653/v1/N18-1013} {Improving implicit
  discourse relation classification by modeling inter-dependencies of discourse
  units in a paragraph}.
\newblock In \emph{Proceedings of the 2018 Conference of the North {A}merican
  Chapter of the Association for Computational Linguistics: Human Language
  Technologies, Volume 1 (Long Papers)}, pages 141--151, New Orleans,
  Louisiana. Association for Computational Linguistics.

\bibitem[{Das and Taboada(2014)}]{das2014rstsc-manual}
Debopam Das and Maite Taboada. 2014.
\newblock {RST} {S}ignalling {C}orpus {A}nnotation {M}anual.
\newblock Technical report, Simon Fraser University.

\bibitem[{Das and Taboada(2017)}]{DasTaboada2017}
Debopam Das and Maite Taboada. 2017.
\newblock Signalling of coherence relations in discourse, beyond discourse
  markers.
\newblock \emph{Discourse Processes}, 55(8):743--770.

\bibitem[{Das et~al.(2019)Das, Taboada, and
  McFetridge}]{DasTaboadaMcFetridge2019}
Debopam Das, Maite Taboada, and Paul McFetridge. 2019.
\newblock {RST Signalling Corpus. LDC2015T10}.

\bibitem[{Demberg et~al.(2019)Demberg, Asr, and Scholman}]{DembergEtAl2019}
Vera Demberg, Fatemeh~Torabi Asr, and Merel Scholman. 2019.
\newblock How compatible are our discourse annotation frameworks? insights from
  mapping {RST-DT} and {PDTB} annotations.
\newblock \emph{Dialogue \& Discourse}, 10(1):87--–135.

\bibitem[{Gessler et~al.(2022)Gessler, Levine, and
  Zeldes}]{gessler-etal-2022-midas}
Luke Gessler, Lauren Levine, and Amir Zeldes. 2022.
\newblock \href {https://aclanthology.org/2022.law-1.13} {{M}idas loop: A
  prioritized human-in-the-loop annotation for large scale multilayer data}.
\newblock In \emph{Proceedings of the 16th Linguistic Annotation Workshop
  (LAW-XVI) within LREC2022}, pages 103--110, Marseille, France. European
  Language Resources Association.

\bibitem[{Gessler et~al.(2019)Gessler, Liu, and
  Zeldes}]{gessler-etal-2019-discourse}
Luke Gessler, Yang Liu, and Amir Zeldes. 2019.
\newblock \href {https://doi.org/10.18653/v1/W19-2708} {A discourse signal
  annotation system for {RST} trees}.
\newblock In \emph{Proceedings of the Workshop on Discourse Relation Parsing
  and Treebanking 2019}, pages 56--61, Minneapolis, MN. Association for
  Computational Linguistics.

\bibitem[{Gessler et~al.(2020)Gessler, Peng, Liu, Zhu, Behzad, and
  Zeldes}]{gessler-etal-2020-amalgum}
Luke Gessler, Siyao Peng, Yang Liu, Yilun Zhu, Shabnam Behzad, and Amir Zeldes.
  2020.
\newblock \href {https://aclanthology.org/2020.lrec-1.648} {{AMALGUM} {--} a
  free, balanced, multilayer {E}nglish web corpus}.
\newblock In \emph{Proceedings of the Twelfth Language Resources and Evaluation
  Conference}, pages 5267--5275, Marseille, France. European Language Resources
  Association.

\bibitem[{Guz and Carenini(2020)}]{guz-carenini-2020-coreference}
Grigorii Guz and Giuseppe Carenini. 2020.
\newblock \href {https://doi.org/10.18653/v1/2020.codi-1.17} {Coreference for
  discourse parsing: A neural approach}.
\newblock In \emph{Proceedings of the First Workshop on Computational
  Approaches to Discourse}, pages 160--167, Online. Association for
  Computational Linguistics.

\bibitem[{Joty et~al.(2013)Joty, Carenini, Ng, and
  Mehdad}]{joty-etal-2013-combining}
Shafiq Joty, Giuseppe Carenini, Raymond Ng, and Yashar Mehdad. 2013.
\newblock \href {https://aclanthology.org/P13-1048} {Combining intra- and
  multi-sentential rhetorical parsing for document-level discourse analysis}.
\newblock In \emph{Proceedings of the 51st Annual Meeting of the Association
  for Computational Linguistics (Volume 1: Long Papers)}, pages 486--496,
  Sofia, Bulgaria. Association for Computational Linguistics.

\bibitem[{Joty et~al.(2015)Joty, Carenini, and Ng}]{joty-etal-2015-codra}
Shafiq Joty, Giuseppe Carenini, and Raymond~T. Ng. 2015.
\newblock \href {https://doi.org/10.1162/COLI_a_00226} {{CODRA}: A novel
  discriminative framework for rhetorical analysis}.
\newblock \emph{Computational Linguistics}, 41(3):385--435.

\bibitem[{Knaebel(2021)}]{knaebel-2021-discopy}
Ren{\'e} Knaebel. 2021.
\newblock \href {https://doi.org/10.18653/v1/2021.codi-main.12} {discopy: A
  neural system for shallow discourse parsing}.
\newblock In \emph{Proceedings of the 2nd Workshop on Computational Approaches
  to Discourse}, pages 128--133, Punta Cana, Dominican Republic and Online.
  Association for Computational Linguistics.

\bibitem[{Kobayashi et~al.(2020)Kobayashi, Hirao, Kamigaito, Okumura, and
  Nagata}]{Kobayashi2020TopDownRP}
Naoki Kobayashi, Tsutomu Hirao, Hidetaka Kamigaito, Manabu Okumura, and Masaaki
  Nagata. 2020.
\newblock \href {https://ojs.aaai.org/index.php/AAAI/article/view/6321/6177}
  {Top-down {RST} parsing utilizing granularity levels in documents}.
\newblock In \emph{AAAI Conference on Artificial Intelligence}.

\bibitem[{Li et~al.(2014)Li, Wang, Cao, and Li}]{li-etal-2014-text}
Sujian Li, Liang Wang, Ziqiang Cao, and Wenjie Li. 2014.
\newblock \href {https://doi.org/10.3115/v1/P14-1003} {Text-level discourse
  dependency parsing}.
\newblock In \emph{Proceedings of the 52nd Annual Meeting of the Association
  for Computational Linguistics (Volume 1: Long Papers)}, pages 25--35,
  Baltimore, Maryland. Association for Computational Linguistics.

\bibitem[{Lin and Zeldes(2021)}]{lin-zeldes-2021-wikigum}
Jessica Lin and Amir Zeldes. 2021.
\newblock \href {https://doi.org/10.18653/v1/2021.law-1.18} {{W}iki{GUM}:
  Exhaustive entity linking for wikification in 12 genres}.
\newblock In \emph{Proceedings of the Joint 15th Linguistic Annotation Workshop
  (LAW) and 3rd Designing Meaning Representations (DMR) Workshop}, pages
  170--175, Punta Cana, Dominican Republic. Association for Computational
  Linguistics.

\bibitem[{Lin et~al.(2019)Lin, Joty, Jwalapuram, and
  Bari}]{lin-etal-2019-unified}
Xiang Lin, Shafiq Joty, Prathyusha Jwalapuram, and M~Saiful Bari. 2019.
\newblock \href {https://doi.org/10.18653/v1/P19-1410} {A unified linear-time
  framework for sentence-level discourse parsing}.
\newblock In \emph{Proceedings of the 57th Annual Meeting of the Association
  for Computational Linguistics}, pages 4190--4200, Florence, Italy.
  Association for Computational Linguistics.

\bibitem[{Liu(2019)}]{liu-2019-beyond}
Yang Liu. 2019.
\newblock \href {https://doi.org/10.18653/v1/W19-2710} {Beyond the {W}all
  {S}treet {J}ournal: Anchoring and comparing discourse signals across genres}.
\newblock In \emph{Proceedings of the Workshop on Discourse Relation Parsing
  and Treebanking 2019}, pages 72--81, Minneapolis, MN. Association for
  Computational Linguistics.

\bibitem[{Liu and Lapata(2017)}]{liu-lapata-2017-learning}
Yang Liu and Mirella Lapata. 2017.
\newblock \href {https://doi.org/10.18653/v1/D17-1133} {Learning contextually
  informed representations for linear-time discourse parsing}.
\newblock In \emph{Proceedings of the 2017 Conference on Empirical Methods in
  Natural Language Processing}, pages 1289--1298, Copenhagen, Denmark.
  Association for Computational Linguistics.

\bibitem[{Liu and Zeldes(2019)}]{liu2019discourse}
Yang Liu and Amir Zeldes. 2019.
\newblock Discourse relations and signaling information: Anchoring discourse
  signals in {RST-DT}.
\newblock \emph{Proceedings of the Society for Computation in Linguistics},
  2(35):314--317.

\bibitem[{Liu and Zeldes(2023)}]{liu-zeldes-2023-eacl}
Yang~Janet Liu and Amir Zeldes. 2023.
\newblock \href {https://aclanthology.org/2023.eacl-main.227} {Why can{'}t
  discourse parsing generalize? {A} thorough investigation of the impact of
  data diversity}.
\newblock In \emph{Proceedings of the 17th Conference of the European Chapter
  of the Association for Computational Linguistics}, pages 3104--3122,
  Dubrovnik, Croatia. Association for Computational Linguistics.

\bibitem[{Liu et~al.(2021)Liu, Shi, and Chen}]{liu-etal-2021-dmrst}
Zhengyuan Liu, Ke~Shi, and Nancy Chen. 2021.
\newblock \href {https://doi.org/10.18653/v1/2021.codi-main.15} {{DMRST}: A
  joint framework for document-level multilingual {RST} discourse segmentation
  and parsing}.
\newblock In \emph{Proceedings of the 2nd Workshop on Computational Approaches
  to Discourse}, pages 154--164, Punta Cana, Dominican Republic and Online.
  Association for Computational Linguistics.

\bibitem[{Mann and Thompson(1988)}]{mann1988rhetorical}
William~C Mann and Sandra~A Thompson. 1988.
\newblock Rhetorical {S}tructure {T}heory: Toward a {F}unctional {T}heory of
  {T}ext {O}rganization.
\newblock \emph{Text-Interdisciplinary Journal for the Study of Discourse},
  8(3):243--281.

\bibitem[{Marcus et~al.(1993)Marcus, Santorini, and
  Marcinkiewicz}]{MarcusSantoriniMarcinkiewicz1993}
Mitchell~P. Marcus, Beatrice Santorini, and Mary~Ann Marcinkiewicz. 1993.
\newblock Building a {L}arge {A}nnotated {C}orpus of {E}nglish: The {P}enn
  {T}reebank.
\newblock \emph{Special Issue on Using Large Corpora, Computational
  Linguistics}, 19(2):313--330.

\bibitem[{Morey et~al.(2017)Morey, Muller, and Asher}]{morey-etal-2017-much}
Mathieu Morey, Philippe Muller, and Nicholas Asher. 2017.
\newblock \href {https://doi.org/10.18653/v1/D17-1136} {How much progress have
  we made on {RST} discourse parsing? a replication study of recent results on
  the {RST}-{DT}}.
\newblock In \emph{Proceedings of the 2017 Conference on Empirical Methods in
  Natural Language Processing}, pages 1319--1324, Copenhagen, Denmark.
  Association for Computational Linguistics.

\bibitem[{Muller et~al.(2012)Muller, Afantenos, Denis, and
  Asher}]{muller-etal-2012-constrained}
Philippe Muller, Stergos Afantenos, Pascal Denis, and Nicholas Asher. 2012.
\newblock \href {https://aclanthology.org/C12-1115} {Constrained decoding for
  text-level discourse parsing}.
\newblock In \emph{Proceedings of {COLING} 2012}, pages 1883--1900, Mumbai,
  India. The COLING 2012 Organizing Committee.

\bibitem[{Nguyen et~al.(2021)Nguyen, Nguyen, Joty, and
  Li}]{nguyen-etal-2021-rst}
Thanh-Tung Nguyen, Xuan-Phi Nguyen, Shafiq Joty, and Xiaoli Li. 2021.
\newblock \href {https://doi.org/10.18653/v1/2021.naacl-main.128} {{RST}
  parsing from scratch}.
\newblock In \emph{Proceedings of the 2021 Conference of the North American
  Chapter of the Association for Computational Linguistics: Human Language
  Technologies}, pages 1613--1625, Online. Association for Computational
  Linguistics.

\bibitem[{Nishida and Nakayama(2020)}]{nishida-nakayama-2020-unsupervised}
Noriki Nishida and Hideki Nakayama. 2020.
\newblock \href {https://doi.org/10.1162/tacl_a_00312} {Unsupervised discourse
  constituency parsing using {V}iterbi {EM}}.
\newblock \emph{Transactions of the Association for Computational Linguistics},
  8:215--230.

\bibitem[{Prasad et~al.(2008)Prasad, Dinesh, Lee, Miltsakaki, Robaldo, Joshi,
  and Webber}]{prasad-etal-2008-penn}
Rashmi Prasad, Nikhil Dinesh, Alan Lee, Eleni Miltsakaki, Livio Robaldo,
  Aravind Joshi, and Bonnie Webber. 2008.
\newblock \href
  {http://www.lrec-conf.org/proceedings/lrec2008/pdf/754_paper.pdf} {The {P}enn
  {D}iscourse {T}ree{B}ank 2.0.}
\newblock In \emph{Proceedings of the Sixth International Conference on
  Language Resources and Evaluation ({LREC}'08)}, Marrakech, Morocco. European
  Language Resources Association (ELRA).

\bibitem[{Prasad et~al.(2014)Prasad, Webber, and
  Joshi}]{prasad-etal-2014-reflections}
Rashmi Prasad, Bonnie Webber, and Aravind Joshi. 2014.
\newblock \href {https://doi.org/10.1162/COLI_a_00204} {Reflections on the
  {P}enn {D}iscourse {T}ree{B}ank, comparable corpora, and complementary
  annotation}.
\newblock \emph{Computational Linguistics}, 40(4):921--950.

\bibitem[{Pyatkin et~al.(2023)Pyatkin, Yung, Scholman, Tsarfaty, Dagan, and
  Demberg}]{pyatkin2023design}
Valentina Pyatkin, Frances Yung, Merel C.~J. Scholman, Reut Tsarfaty, Ido
  Dagan, and Vera Demberg. 2023.
\newblock \href {http://arxiv.org/abs/2304.00815} {Design choices for
  crowdsourcing implicit discourse relations: Revealing the biases introduced
  by task design}.

\bibitem[{Scholman et~al.(2022)Scholman, Pyatkin, Yung, Dagan, Tsarfaty, and
  Demberg}]{scholman-etal-2022-design}
Merel Scholman, Valentina Pyatkin, Frances Yung, Ido Dagan, Reut Tsarfaty, and
  Vera Demberg. 2022.
\newblock \href {https://aclanthology.org/2022.lrec-1.231} {Design choices in
  crowdsourcing discourse relation annotations: The effect of worker selection
  and training}.
\newblock In \emph{Proceedings of the Thirteenth Language Resources and
  Evaluation Conference}, pages 2148--2156, Marseille, France. European
  Language Resources Association.

\bibitem[{Shang et~al.(2019)Shang, Liu, Wang, Rong, and Liu}]{ShangEtAl2019}
Erbo Shang, Xiaohua Liu, Hailong Wang, Yangfeng Rong, and Yuerong Liu. 2019.
\newblock \href {https://doi.org/10.1109/IAEAC47372.2019.8997992} {Research on
  the application of artificial intelligence and distributed parallel computing
  in archives classification}.
\newblock In \emph{2019 IEEE 4th Advanced Information Technology, Electronic
  and Automation Control Conference (IAEAC)}, pages 1267--1271.

\bibitem[{Sporleder and Lascarides(2004)}]{sporleder-lascarides-2004-combining}
Caroline Sporleder and Alex Lascarides. 2004.
\newblock \href {https://aclanthology.org/C04-1007} {Combining hierarchical
  clustering and machine learning to predict high-level discourse structure}.
\newblock In \emph{{COLING} 2004: Proceedings of the 20th International
  Conference on Computational Linguistics}, pages 43--49, Geneva, Switzerland.
  COLING.

\bibitem[{Stede and Neumann(2014)}]{StedeNeumann2014}
Manfred Stede and Arne Neumann. 2014.
\newblock {Potsdam Commentary Corpus} 2.0: Annotation for discourse research.
\newblock In \emph{Proceedings of the Language Resources and Evaluation
  Conference (LREC '14)}, pages 925--929, Reykjavik.

\bibitem[{Wang et~al.(2017)Wang, Li, and Wang}]{wang-etal-2017-two}
Yizhong Wang, Sujian Li, and Houfeng Wang. 2017.
\newblock \href {https://doi.org/10.18653/v1/P17-2029} {A two-stage parsing
  method for text-level discourse analysis}.
\newblock In \emph{Proceedings of the 55th Annual Meeting of the Association
  for Computational Linguistics (Volume 2: Short Papers)}, pages 184--188,
  Vancouver, Canada. Association for Computational Linguistics.

\bibitem[{Webber et~al.(2019)Webber, Prasad, Lee, and
  Joshi}]{PDTB3-Annotation-Manual}
Bonnie Webber, Rashmi Prasad, Alan Lee, and Aravind Joshi. 2019.
\newblock The {Penn Discourse TreeBank} 3.0 annotation manual.
\newblock Technical report, University of Edinburgh, Interactions, LLC,
  University of Pennsylvania.

\bibitem[{Yu et~al.(2018)Yu, Zhang, and Fu}]{yu-etal-2018-transition}
Nan Yu, Meishan Zhang, and Guohong Fu. 2018.
\newblock \href {https://aclanthology.org/C18-1047} {Transition-based neural
  {RST} parsing with implicit syntax features}.
\newblock In \emph{Proceedings of the 27th International Conference on
  Computational Linguistics}, pages 559--570, Santa Fe, New Mexico, USA.
  Association for Computational Linguistics.

\bibitem[{Yu et~al.(2022)Yu, Zhang, Fu, and Zhang}]{yu-etal-2022-rst}
Nan Yu, Meishan Zhang, Guohong Fu, and Min Zhang. 2022.
\newblock \href {https://doi.org/10.18653/v1/2022.acl-long.294} {{RST}
  discourse parsing with second-stage {EDU}-level pre-training}.
\newblock In \emph{Proceedings of the 60th Annual Meeting of the Association
  for Computational Linguistics (Volume 1: Long Papers)}, pages 4269--4280,
  Dublin, Ireland. Association for Computational Linguistics.

\bibitem[{Zeldes(2017)}]{Zeldes2017}
Amir Zeldes. 2017.
\newblock \href {https://doi.org/http://dx.doi.org/10.1007/s10579-016-9343-x}
  {The {GUM} {C}orpus: Creating multilayer resources in the classroom}.
\newblock \emph{Language Resources and Evaluation}, 51(3):581--612.

\bibitem[{Zeldes et~al.(2021)Zeldes, Liu, Iruskieta, Muller, Braud, and
  Badene}]{zeldes-etal-2021-disrpt}
Amir Zeldes, Yang~Janet Liu, Mikel Iruskieta, Philippe Muller, Chlo{\'e} Braud,
  and Sonia Badene. 2021.
\newblock \href {https://doi.org/10.18653/v1/2021.disrpt-1.1} {The {DISRPT}
  2021 shared task on elementary discourse unit segmentation, connective
  detection, and relation classification}.
\newblock In \emph{Proceedings of the 2nd Shared Task on Discourse Relation
  Parsing and Treebanking (DISRPT 2021)}, pages 1--12, Punta Cana, Dominican
  Republic. Association for Computational Linguistics.

\bibitem[{Zhao and Webber(2021)}]{zhao-webber-2021-revisiting}
Zheng Zhao and Bonnie Webber. 2021.
\newblock \href {https://doi.org/10.18653/v1/2021.codi-main.10} {Revisiting
  shallow discourse parsing in the {PDTB}-3: Handling intra-sentential
  implicits}.
\newblock In \emph{Proceedings of the 2nd Workshop on Computational Approaches
  to Discourse}, pages 107--121, Punta Cana, Dominican Republic and Online.
  Association for Computational Linguistics.

\bibitem[{Zhu et~al.(2021)Zhu, Pradhan, and Zeldes}]{zhu-etal-2021-ontogum}
Yilun Zhu, Sameer Pradhan, and Amir Zeldes. 2021.
\newblock \href {https://doi.org/10.18653/v1/2021.acl-short.59} {{O}nto{GUM}:
  Evaluating contextualized {SOTA} coreference resolution on 12 more genres}.
\newblock In \emph{Proceedings of the 59th Annual Meeting of the Association
  for Computational Linguistics and the 11th International Joint Conference on
  Natural Language Processing (Volume 2: Short Papers)}, pages 461--467,
  Online. Association for Computational Linguistics.

\end{thebibliography}
\bibliographystyle{acl_natbib}

\appendix


\section{Validation Performance}
\label{appendix:training-and-val-performance}

Table \ref{tab:parser-results-dev} shows our reproduced 5-run average parsing performance on the \texttt{dev} partition of each corpus. GUM v9 has an established \texttt{dev} partition following the UD English GUM treebank. While RST-DT does not have an established \texttt{dev} partition, we followed previous work by taking $10$\% of training data stratified by the number of EDUs in each document \cite{guz-carenini-2020-coreference}, which remained the same in the training for both parsers. 
The list of document names used as development data can be found in the repository of the paper for reproducibility purposes. 

\begin{table}[ht]
\resizebox{\columnwidth}{!}{%
\begin{tabular}{@{}l|ccc|ccc@{}}
\toprule
\textit{corpora}                                        & \multicolumn{3}{c|}{\textbf{GUM v9}} & \multicolumn{3}{c}{\textbf{RST-DT}}  \\ \midrule
\textit{metrics}                                        & \textbf{S} & \textbf{N} & \textbf{R} & \textbf{S} & \textbf{N} & \textbf{R} \\ \midrule
\begin{tabular}[c]{@{}l@{}}\textsc{bottom-up}\\ \citet{guz-carenini-2020-coreference}\end{tabular} & $67.9$       & $64.8 $      & $46.8$       & $76.0$       & $64.9$       & $55.2$       \\ \midrule
\begin{tabular}[c]{@{}l@{}}\textsc{top-down}\\ \citet{liu-etal-2021-dmrst}\end{tabular}  & $69.3$       & $56.3$       & $48.1$       & $75.0$       & $64.6$       & $55.7 $      \\ \bottomrule
\end{tabular}%
}
\caption{Validation Performance on GUM v9 and RST-DT with Gold Segmentation (5 run average).}
\label{tab:parser-results-dev}
\end{table}

\end{document}